\documentclass[sigconf,nonacm]{acmart}

\usepackage[section]{placeins}

\AtBeginDocument{%
  \providecommand\BibTeX{{%
    \normalfont B\kern-0.5em{\scshape i\kern-0.25em b}\kern-0.8em\TeX}}}

\begin{document}

\title{Runtime phylogenetic analysis enables extreme subsampling for test-based problems}

\author{Alexander Lalejini}
\affiliation{%
  \institution{Grand Valley State University}
  \country{}
}
\email{lalejina@gvsu.edu}

\author{Marcos Sanson}
\affiliation{%
  \institution{Grand Valley State University}
  \country{}
}
\email{sansonm@mail.gvsu.edu}

\author{Jack Garbus}
\affiliation{%
  \institution{Brandeis University}
  \country{}
}
\email{garbus@brandeis.edu}

\author{Matthew Andres Moreno}
\affiliation{%
  \institution{University of Michigan}
  \country{}
}
\email{morenoma@umich.edu}

\author{Emily Dolson}
\affiliation{%
  \institution{Michigan State University}
  \country{}
}
\email{dolsonem@msu.edu}

\renewcommand{\shortauthors}{Lalejini et al.}

\begin{abstract}

A phylogeny describes the evolutionary history of an evolving population.
Evolutionary search algorithms can perfectly track the ancestry of candidate solutions, illuminating a population's trajectory through the search space.
However, phylogenetic analyses are typically limited to post-hoc studies of search performance.
We introduce phylogeny-informed subsampling, a new class of subsampling methods that exploit runtime phylogenetic analyses for solving test-based problems.
Specifically, we assess two phylogeny-informed subsampling methods---individualized random subsampling and ancestor-based subsampling---on three diagnostic problems and ten genetic programming (GP) problems from program synthesis benchmark suites.
Overall, we found that phylogeny-informed subsampling methods enable problem-solving success at extreme subsampling levels where other subsampling methods fail.
For example, phylogeny-informed subsampling methods more reliably solved program synthesis problems when evaluating just one training case per-individual, per-generation.
However, at moderate subsampling levels, phylogeny-informed subsampling generally performed no better than random subsampling on GP problems.
Our diagnostic experiments show that phylogeny-informed subsampling improves diversity maintenance relative to random subsampling, but its effects on a selection scheme's capacity to rapidly exploit fitness gradients varied by selection scheme.
Continued refinements of phylogeny-informed subsampling techniques offer a promising new direction for scaling up evolutionary systems to handle problems with many expensive-to-evaluate fitness criteria.

\end{abstract}



\maketitle

\section{Introduction}

A phylogeny (ancestry tree) describes the evolutionary history of an evolving population by representing the evolutionary relationships among taxa (e.g., individuals, genotypes, genes, species, \textit{etc.}). 
Evolutionary biologists typically estimate phylogenies from the fossil record, phenotypic traits, and extant genetic information.
Phylogenetic analyses using such imperfect phylogenies have profoundly advanced our understanding of life on Earth \citep{raupExtinctionBadGenes1992, wileyPhylogeneticsTheoryPractice2011,rothan_epidemiology_2020,tucker_guide_2017}. 
Evolutionary search algorithms, in contrast, can track the ancestry of candidate solutions with perfect (or user adjustable) accuracy~\citep{ofria_empirical_2020,moreno_hstrat_2022,de_rainville_deap_2012,bohm_mabe_2017,godin-dubois_apoget_2019,dolson_phylotrackpy_2023}.
Indeed, post-hoc phylogenetic analyses have yielded valuable insights for evolutionary computing research \citep{mcpheeUsingGraphDatabases2016,hernandezWhatCanPhylogenetic2022, shahbandeganUntanglingPhylogeneticDiversity2022,dolson_interpreting_2020,mcphee_analysis_1999,donatucci_analysis_2014,burlacu_population_2023,mcphee_visualizing_2017}, but phylogenetic analyses remain largely unexplored for assisting evolutionary search at runtime.
In this work, we introduce phylogeny-informed subsampling methods that exploit phylogenetic analyses for solving test-based problems. 

In evolutionary computing, many problems require evaluating candidate solutions on many fitness criteria to accurately assess solution quality.
This scenario is especially common in genetic programming (GP) where, each generation, a population of evolving programs is typically evaluated on a corpus of input-output examples (i.e., training cases) that specify expected program behavior.
Using the results of these training case evaluations, a selection algorithm chooses promising programs to ``reproduce'' into the next generation.
Ideally, we would thoroughly evaluate each candidate solution's quality with a large number of training cases.
However, thoroughly assessing candidate solution quality can be an inefficient use of computational resources, especially if training cases (or fitness criteria) are computationally expensive to evaluate or assess redundant behavior~\citep{boldi_informed_2023}.
Finite computing resources force GP systems to balance evaluation thoroughness with other aspects of evolutionary search, such as population size or generation count, which are also important for problem-solving success \citep{hernandezRandomSubsamplingImproves2019, fergusonCharacterizingEffectsRandom2020, helmuthProblemSolvingBenefitsDownSampled2022,helmuth_benchmarking_2020,briesch_trade-off_2023}.  

Evaluation thoroughness is often traded for computational savings on test-based problems by assessing each candidate solution on a small subsample of training cases each generation~\citep{jin_comprehensive_2005,jin_surrogate-assisted_2011,kanade_towards_2004,goos_dynamic_1994}.
Indeed, random subsampling has been shown to dramatically improve problem-solving success in GP by reallocating computational savings to run search for more generations or to increase population size~\citep{fergusonCharacterizingEffectsRandom2020,helmuthProblemSolvingBenefitsDownSampled2022,geiger_down-sampled_2023}. 
Current subsampling methods, however, sometimes omit important evaluation criteria, potentially causing the loss of genetic diversity and negatively affecting problem-solving success~\citep{hernandez_exploration_2022,boldi_static_2023,fergusonCharacterizingEffectsRandom2020}.
Moderating subsample rates or using many redundant training cases can mitigate this drawback. 
Unfortunately, even the occasional loss of critical diversity from a population can slow or prevent problem-solving success on certain problems. 
This risk is particularly pronounced on problems that require substantial search space exploration or those where an expected behavior is represented by only a small proportion of training cases (making them more likely to be left out in a random subsample)~\citep{lozano_genetic_diversity_2008}. 
Some subsampling techniques (e.g., informed down-sampling~\citep{boldi_informed_2023}) mitigate these issues by using population statistics at runtime to try to build subsamples that better represent the full training set.

As phylogenies are a rich source of information about the process of evolution, we hypothesize that we can build on the success of techniques like informed down-sampling~\citep{boldi_informed_2023} by using phylogenetic information.
We propose a new category of phylogeny-informed subsampling procedures for test-based problems.
These methods operate on the assumption that the population's ancestry information is tracked during evolution. 
In addition to tracking ancestral relationships, the phylogeny is annotated with any training case results evaluated during a run. 
Ancestral performance data can then be used to (1) \textit{estimate} a candidate solution's performance on unevaluated training cases  and (2) \textit{choose} which training cases should be included in a subsample. 
Previous work demonstrated that phylogeny-based \textit{estimations} used in combination with random down-sampling can improve diversity maintenance and overall problem-solving success at extreme subsampling rates~\citep{lalejini_phylo_est_2023}.
Here, we extend the use of phylogenetic analysis to choose \textit{individualized} subsamples when evaluating candidate solutions. 

We introduce two phylogeny-informed subsampling routines in the context of lexicase parent selection: individualized random subsampling (IRS) and ancestor-based subsampling (ABS).
Each generation, IRS constructs a random subsample of training cases for each candidate solution to be evaluated on and uses phylogenetic information to estimate the solution's performance on all other cases. 
Like IRS, ABS constructs a subsample of training cases for each candidate solution. 
However, instead of selecting cases randomly, ABS chooses the training cases on which the candidate solution's ancestors were least recently evaluated, minimizing the phylogenetic distance from which estimations are drawn. 
We compared IRS, ABS, down-sampling with phylogeny-based fitness estimation~\citep{lalejini_phylo_est_2023}, and down-sampling with no estimation on thirteen problems across two problem domains: three selection scheme diagnostic problems~\citep{hernandez_suite_2022} and ten program synthesis benchmark problems~\citep{helmuth_general_2015,helmuth_psb2_2021}.

\section{Assisting evolutionary search with phylogenetic analysis}
\label{sec:phylo-subsampling}

Many parent selection algorithms require that each candidate solution be evaluated on a comparable set of training cases each generation in order to assess their relative performance.
Conventional subsampling methods accommodate this constraint by constructing a single sample to evaluate all individuals against. 
For example, random down-sampling chooses a subsample of training cases at random each generation.
Phylogeny-informed subsampling methods allow for \textit{individualized} subsampling; that is, each candidate solution can be evaluated on a unique sample of training cases while remaining comparable to other members of the population.
Phylogeny-informed subsampling maintains comparability among candidate solutions by estimating candidate solution performance on any unevaluated training cases via phylogeny-based fitness estimation~\citep{lalejini_phylo_est_2023}. Thus, any parent selection algorithm can be applied as if all candidates were evaluated against all training cases.

\subsection{Phylogeny-informed fitness estimation}
\label{sec:phylo-subsampling:fitness-est}

\citet{lalejini_phylo_est_2023} recently proposed two phylogeny-informed fitness estimation methods: ancestor-based estimation and relative-based estimation. 
In our experiments, we use ancestor-based estimation, as it is computationally cheaper and neither estimation method was found to be consistently better than the other. 
To estimate an individual's performance on a training case, ancestor-based estimation iterates over the individual's direct ancestors until finding the nearest ancestor that was evaluated against the focal training case. 
This ancestor's score then serves as the estimate for the focal individual's performance on that training case. 

Phylogeny-informed fitness estimation requires the overhead of ancestry tracking.
In our experiments, we use the same tracking approach as in~\citep{lalejini_phylo_est_2023}, which discusses the computational costs in detail. 
Phylogenies can be tracked at any taxonomic level of organization (e.g., individuals, genotypes, \textit{etc}.), and a node in a phylogenetic tree is generically referred to as a ``taxon'' (pluralized as ``taxa'').
In our work, taxa represented genotypes.
We limited the computational cost of estimation by constraining search distance in the phylogeny to eight steps, and we assumed maximally poor performance for any failed estimates (as in~\citep{lalejini_phylo_est_2023}). 

Ancestor-based estimation can be optimized to forgo full phylogeny tracking by passing evaluation information from parent to offspring.
This optimization is similar to many fitness inheritance methods where a proportion of offspring inherit the mean fitness of their parents~\citep{bui_fitness_2005,chen_fitness_2002,fonseca_study_2012,ducheyne_fitness_2008,wang_fitness_2018,shi_asaga_2008}.
\citet{pilato_fitness_2007,hiot_speeding-up_2010}'s ancestor-based fitness inheritance method works similarly to our estimation approach, but uses multiple genetically similar ancestors to compute an estimate. 
Phylogeny-informed fitness estimation extends existing fitness inheritance methods by conducting independent estimates for each training case, supporting the use of collateral relatives for estimates, and by allowing additional phylogenetic information to be incorporated into the estimate if so desired. 
Like fitness inheritance methods, phylogeny-informed estimation assumes that offspring behavior will (on average) be strongly correlated with their parents.

\citet{lalejini_phylo_est_2023} applied phylogeny-informed estimation in combination with random down-sampling and cohort partitioning (each using lexicase selection).
\citet{lalejini_phylo_est_2023} found that estimation could improve diversity maintenance and search space exploration but did not observe consistent benefits for problem-solving success across four GP problems.
Phylogeny-informed subsampling builds on these fitness estimation methods by customizing subsample construction for each member of the population. 

\subsection{Phylogeny-informed subsampling}

In this work, we introduce two simple phylogeny-informed subsampling methods: individualized random subsampling and ancestor-based subsampling. 

\subsubsection{Individualized random sampling (IRS)}
\label{sec:phylo-subsampling:irs}

IRS constructs a random subsample of training cases for each member of the population. 
Candidate solutions are then assessed on their assigned random sample. 
IRS is a straightforward extension of random down-sampling, but has the benefit of using a greater diversity of training cases when assessing a population each generation.
While phylogenetic analysis is not used to construct subsamples, phylogenetic information enables IRS by ensuring that all members of the population are comparable by the selection algorithm.

\subsubsection{Ancestor-based subsampling (ABS)}
\label{sec:phylo-subsampling:abs}

Ancestor-based subsampling (ABS) constructs a subsample based on which training cases a candidate solution's recent ancestors were evaluated against. 
When building a subsample of size $S$, all training cases are initially eligible for inclusion.
We then iterate over the candidate solution's taxon and its direct ancestors in sequence.
At each step, we mark all training cases that the current taxon has already been evaluated against as ineligible.
This process continues until the number of eligible training cases is less than or equal to $S$.
At that point, all remaining eligible training cases are added to the subsample.
To fill out the subsample, random training cases from the set most recently marked ineligible are added until it contains $S$ training cases.
In this way, ABS prioritizes evaluating candidate solutions on training cases least recently evaluated in that candidate's ancestry.

\section{Methods}
\label{sec:methods}

We assessed phylogeny-informed subsampling methods in the context of the lexicase parent selection algorithm on thirteen problems across two problem domains: three selection scheme diagnostic problems~\citep{hernandez_suite_2022} and ten GP problems selected from the first and second program synthesis benchmark suites~\citep{helmuth_general_2015,helmuth_psb2_2021}.
We repeated all experiments at two subsampling rates: 1\% and 10\%. 

\subsection{Lexicase Selection}
\label{sec:methods:lexicase}

Lexicase selection is designed for use on test-based problems where candidate solutions are evaluated on a set of training cases that specify correct behavior \citep{spectorAssessmentProblemModality2012}.
Lexicase selection has been successful across many domains~\citep{aenugu_lexicase_2019,moore_lexicase_2017,lalejini_artificial_2022,ding_optimizing_2022,matsumoto_faster_2023,boldi_improving_2023,la_cava_epsilon_lexicase_2016,metevier_lexicase_2019,moore_lexicase_2022}, including (and especially) genetic programming~\citep{helmuth_benchmarking_2020,orzechowski_where_2018}. 
To select a parent, lexicase selection first shuffles all training cases and adds all members of the population to a pool of candidates eligible for selection. 
Each training cases is then used in sequence (in shuffled order) to remove all but the elite candidates (on the current training case) from the pool of selectable candidates. 
This process continues until all training cases have been used. 
If more than one candidate remains in the pool, one is selected at random. 

\subsubsection{Down-sampled lexicase selection}

Each generation, down-sampled lexicase randomly subsamples the training set used to evaluate and select parents~\citep{hernandezRandomSubsamplingImproves2019}. 
When augmented with phylogeny-informed estimation, the full training set can be used for parent selection, and the population's phylogeny is used to estimate any performances on training cases a candidate solution was not evaluated against.

\subsubsection{Applying phylogeny-informed subsampling to lexicase selection}

Individualized random subsampling and ancestor-based subsampling choose the training cases used to evaluate candidate solutions on an individual basis. 
Once all members of the population are evaluated on their subsample of training cases, the complete training set is used for lexicase selection, and the population's phylogeny is used to estimate performances on any training cases a candidate solution was not evaluated against.

\subsection{Diagnostic experiments}
\label{sec:methods:diagostics}

We used the exploitation rate, contradictory objectives, and multi-path exploration diagnostics from the DOSSIER suite to compare four subsampling regimes: individualized random sampling, ancestor-based sampling, random down-sampling with fitness estimation, and random down-sampling without fitness estimation. 
The DOSSIER suite is a set of diagnostic benchmarks that measure important problem-solving characteristics for evolutionary search~\citep{hernandez_suite_2022}.
These diagnostics have been used to study the problem-solving and phylogenetic characteristics of lexicase selection (among many other selection methods)~\citep{hernandez_exploration_2022,hernandezWhatCanPhylogenetic2022,shahbandeganUntanglingPhylogeneticDiversity2022}, making them an ideal domain to assess the impacts of phylogeny-informed subsampling.
We parameterize our experiments similarly to those from~\citep{hernandez_suite_2022}. 

Each diagnostic represents genomes as a vector of 100 ``genes''.
Genes are floating point values, each between 0.0 and 100.0. 
Each diagnostic specifies a particular genotype-to-phenotype translation that interprets a genome as an equal-length phenotype vector of numeric ``traits''. 
Lexicase selection uses each trait as a training case, so each of our diagnostic experiments use training sets of size 100.
For all diagnostics used in this work, higher trait scores are better. 
We apply subsampling to diagnostic problems in the same way as in~\citep{hernandez_exploration_2022}. 

For each diagnostic experiment, we ran 20 replicates of each condition. 
We evolved populations of size 500 for 50,000 generations. 
Each generation, we evaluated members of the population according to the condition-specific subsampling regime, and we selected parents to reproduce asexually. 
We mutated offspring at a per-gene rate of 0.7\%, drawing mutations from a Gaussian distribution with a mean of 0.0 and standard deviation of 1.0~\citep{hernandez_suite_2022}.

Note that \textit{generations} are held constant across diagnostic experiments regardless of subsampling level, as we limit comparisons to subsampling methods (and not across sampling levels).
We provide a description of relevant diagnostics below but direct readers to~\citep{hernandez_suite_2022} for more detailed descriptions. 

\subsubsection{Exploitation rate diagnostic}

The exploitation rate diagnostic measures a selection scheme's ability to exploit a search space with single, smooth fitness gradient (i.e., hill climbing)~\citep{hernandez_suite_2022}.
This diagnostic directly translates a genome into the phenotype, interpreting each gene value as the corresponding trait value. 
All traits are maximized at the upper bound of 100. 
This creates a search space with a single global optimum comprising a genome where all gene-values are maximized.
For this diagnostic, we report the best aggregate trait score in the population, as greater aggregate trait scores indicate more successful search space exploitation. 

\subsubsection{Contradictory objectives diagnostic}

The contradictory objectives diagnostic measures a selection scheme's ability to find and maintain many independent global optima in a population~\citep{hernandez_suite_2022}. 
To translate a genome into a phenotype, this diagnostic marks the maximum gene as active and all others as inactive. 
The active gene value is copied into the corresponding phenotypic trait, and all traits associated with inactive genes are set to zero.
Thus, a candidate solution's fitness is limited to a single trait value, creating 100 independent global optima in this search space (each associated with a different trait). 
All traits are maximized at the upper bound of 100, and for our analyses, we consider any trait with a value above 98 as ``satisfied''. 
On this diagnostic, we report ``satisfactory trait coverage'', which counts the number of distinct traits satisfied across the entire population. 
Satisfactory trait coverage quantifies the number of unique global optima maintained in a population. 

\subsubsection{Multi-path exploration diagnostic}

The multi-path exploration diagnostic measures a selection scheme's capacity to continuously explore different fitness gradients in a search space~\citep{hernandez_suite_2022}.  
To translate a genome, this diagnostic first marks the maximum gene in the genome as the activation gene. 
Each consecutive gene after the activation gene that is less than or equal to the previous gene is marked as active, creating an active gene region. 
All genes in the active gene region are copied into their corresponding phenotypic traits, and all traits associated with genes outside of the active region are set to zero. 
Traits are maximized at the upper bound of 100, and the phenotype with all maximized traits is the global optimum.  

The multi-path exploration diagnostic defines a search space with many pathways that each begin at a different activation gene. 
Each pathway in the search space has a different path length and peak fitness, but all paths have identical slopes.  
The pathway beginning at the first gene position leads to the global optimum, and all others lead to local optima. 
To be consistently successful on this diagnostic, a selection scheme must be capable of continuously exploring different pathways to reach the global optima.
For this diagnostic, we report the best aggregate trait score in the population because greater scores indicate that a population was able to find and optimize better pathways in the search space.   

\subsection{Genetic programming experiments}
\label{sec:methods:gp}

We compared the problem-solving success of individualized random sampling, ancestor-based sampling, and random down-sampling with and without estimation on ten problems from the first and second general program synthesis benchmark suites~\citep{helmuth_psb2_2021,helmuth_general_2015}:  
Bouncing Balls, Dice Game, Fizz Buzz, For Loop Index, GCD, Grade, Median, Small or Large, Smallest, and Snow Day.

\subsubsection{GP system}

For all GP experiments, we ran 50 replicates of each condition.
In each replicate, we evolved a population of 1,000 linear genetic programs using the SignalGP representation~\citep{lalejini_evolving_2018}.
We configured these experiments similarly to those in~\citep{lalejini_phylo_est_2023}. 
Our supplemental material (see ~\citep{supp_gh_repo}) contains the full instruction sets used and configuration details used for each problem (including source code).
We reproduced programs asexually, applying mutations to offspring. 
We applied single-instruction insertions, deletions, and substitutions, each at a per-instruction rate of 0.1\%. 
We applied single-argument substitutions at a per-argument rate of 0.1\%, and we applied slip mutations~\citep{lalejini_gene_2017}, which can duplicate or delete whole sequences of instructions, at a per-program rate of 5\%.
For all problems, we limited program length to a maximum of 128 instructions, and programs were given 128 time steps (i.e., instruction-execution steps) to return their output. 

\subsubsection{Program synthesis problems}

The program synthesis benchmark suites include a large number of introductory to programming problems that are well studied~\citep{helmuth_general_2015,helmuth_psb2_2021}. 
The problems used in this work range in difficulty from easy (e.g., Smallest) to challenging (e.g., Bouncing Balls)~\citep{helmuth_applying_2022}.
Brief descriptions of each problem are given below: 

\begin{itemize}

\item \textbf{Bouncing Balls}~\citep{helmuth_psb2_2021}: Programs are given three inputs: a starting height (float in [1.0, 100.0]), a height after the first bounce of a dropped ball (float in [1.0 and 100.0]), and a number of bounces (integer in [1 and 20]). 
A program must compute the \textit{bounciness index} using the first two inputs, and then output the total distance that the dropped ball travels across all of its bounces. 

\item \textbf{Dice Game}~\citep{helmuth_psb2_2021}: Programs receive two integers, $n$ and $m$, in range [1, 1000]. 
Each integer corresponds to the number of sides on a die, and  programs must output the probability that an $n$-sided die rolls strictly higher than an $m$-sided die.

\item \textbf{Fizz Buzz}~\citep{helmuth_psb2_2021}: Given an integer $x$, a program must output ``Fizz'' if $x$ is divisible by 3, ``Buzz'' if $x$ is divisible by 5, ``FizzBuzz'' if $x$ is divisible by both 3 and 5, and $x$ if none of those conditions are true.

\item \textbf{For Loop Index}~\citep{helmuth_general_2015}: Programs receive three integer inputs: \textit{start} (in range [-500, 500]), \textit{end} (in range [-500, 500]), and \textit{step} (in range [1, 10]). Beginning with \textit{start}, programs output each subsequent value \textit{step} away from the previous value until \textit{end}.

\item \textbf{GCD}~\citep{helmuth_psb2_2021}: Programs are given two integers (in range [1, 1000000]) and must output the largest integer that evenly divides into each. 

\item \textbf{Grade}~\citep{helmuth_general_2015}: Programs receive five integer inputs in range [0, 100]: $A$, $B$, $C$, $D$, and \textit{score}. 
$A$, $B$, $C$, and $D$ are monotonically decreasing and unique ``grade'' thresholds that describe the minimum score to receive that grade.
A program must return the appropriate letter grade for the given \textit{score} or return $F$ if $\text{\textit{score}} < D$. 

\item \textbf{Median}~\citep{helmuth_general_2015}: Programs receive three integer inputs in range [-100, 100] and must output the median value. 

\item \textbf{Small or Large}~\citep{helmuth_general_2015}: Given an integer $n$, programs must output ``small'' if $n < 1000$, ``large'' if $n \geq 2000$, and ``neither'' if $1000 \geq n < 2000$.

\item \textbf{Smallest}~\citep{helmuth_general_2015}: Programs receive four integer inputs in range [-100, 100] and must output the smallest value. 

\item \textbf{Snow Day}~\citep{helmuth_psb2_2021}: Programs receive four inputs: a number of hours (integer in [0, 20]), how much snow is on the ground (float in  [0.0, 20.0]), the rate of snow fall (float in [0.0, 10.0]), and the proportion of snow melting per hour (float in [0.0, 1.0]). 
Programs must output the amount of snow on the ground after the given number of hours. 

\end{itemize}

For each problem, we used a set of 100 training cases for program evaluation and a set of 1,000 testing cases for determination of problem-solving success. 
All training and testing sets were generated from data provided by the benchmark suite authors ~\cite{helmuth_general_2015,helmuth_psb2_2021}. 
We provide the training and testing sets used for each problem in our supplemental material~\citep{supp_gh_repo}. 
We categorized a replicate as successful if it produced a program that solved all testing cases. 
For all problems, we ensured that input-output edge cases were included in the training and testing sets to ensure that testing sets adequately assessed generalization on their own.
We terminated runs after 50,000,000 training case evaluations, which corresponds to 500 generations of evolution under standard lexicase selection, 5,000 generations under 10\% subsampling, and 50,000 generations under 1\% subsampling.

\subsection{Statistical analyses}

For all experiments, we only made comparisons among different subsampling methods.
We did not compare measurements taken from treatments across different problems or subsampling levels (1\% and 10\%).
When comparing distributions of measurements taken from different treatments, we performed Kruskal-Wallis tests to screen for statistical differences among independent conditions.
For comparisons in which the Kruskal-Wallis test was significant (significance level of 0.05), we performed post-hoc Wilcoxon rank-sum tests to identify pairwise differences. 
When comparing problem-solving success rates, we used pairwise Fisher's exact tests (significance level of 0.05). 
We used the Holm-Bonferroni method to correct for multiple comparisons as appropriate. 
Complete results of statistical analyses are available in our supplemental material~\citep{supp_gh_repo}. 

\subsection{Software and data availability}

Experiment software and data analyses can be found in our supplemental material, which is hosted on GitHub and archived on Zenodo: \citep{supp_gh_repo}.
Our experiment data are archived on the Open Science Framework at~\citep{supp_osf_data}.

\section{Results and discussion}

\subsection{Random down-sampling without fitness estimation most improves gradient exploitation with lexicase selection}

\begin{figure}[h]
\begin{center}
\includegraphics[width=\linewidth]{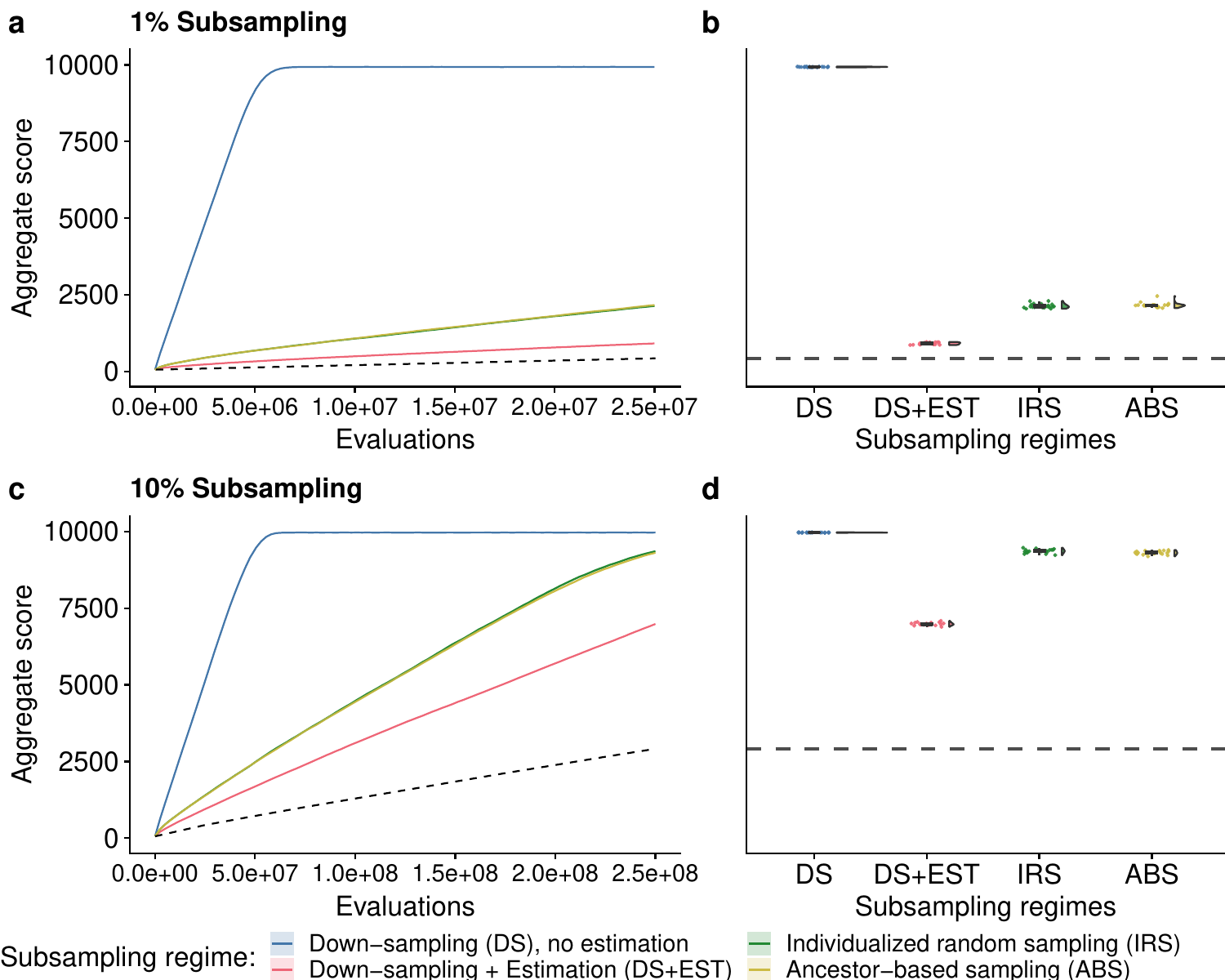}
\caption{\small{
\textbf{Aggregate trait score on the exploitation-rate diagnostic for subsampling regimes with lexicase selection.}
Panels (a) and (c) show mean aggregate trait score over time for 1\% and 10\% subsampling levels, respectively.
Shading around the mean indicates a bootstraped 95\% confidence interval. 
Panels (b) and (d) show best aggregate trait score after 50,000 generations of evolution for 1\% and 10\% subsampling levels, respectively.  
Dotted black lines in each plot indicate the median aggregate trait score achieved by standard lexicase selection (no subsampling) after an equivalent number of trait evaluations.
Kruskal-Wallis tests for both subsampling levels were statistically significant ($p < 0.001$). 
}}
\label{fig:exploit-lex}
\end{center}
\end{figure}

The exploitation rate diagnostic measures a selection scheme's ability to rapidly guide a population up a smooth, unimodal fitness gradient. 
Figure~\ref{fig:exploit-lex} shows aggregate scores over time and after 50,000 generations of evolution under both 1\% and 10\% subsampling rates. 
The dotted black line on each graph indicates the median aggregate score achieved by 20 replicates of standard lexicase selection (with no subsampling) after an equivalent number of trait \textit{evaluations} (e.g., 50,000 generations at 1\% subsampling is equivalent to 500 generations of standard lexicase).
Standard lexicase's median performance is provided for reference, but we did not make direct comparisons to it.

At both 1\% and 10\% subsampling levels, random down-sampling with no estimation (DS) substantially outperforms all other subsampling regimes (corrected Wilcoxon rank-sum tests, $p < 0.001$).
Both phylogeny-informed subsampling regimes, individualized random sampling (IRS) and ancestor-based subsampling (ABS), outperformed random down-sampling with fitness estimation (DS+EST) for both 1\% and 10\% subsampling levels ($p < 0.001$). 
All subsampling regimes achieved greater aggregate scores than standard lexicase run for an equivalent number of trait evaluations. 

Previous work benchmarked standard lexicase selection on the exploitation rate diagnostic~\citep{hernandez_suite_2022}, establishing that lexicase optimizes populations more slowly than elitist aggregative selection methods that do not actively maintain diverse specialists. 
For example, tournament selection and truncation selection performed substantially better than lexicase selection on this diagnostic~\citep{hernandez_suite_2022}. 
Lexicase selection, however, exploited faster than other selection methods with mechanisms for diversity maintenance (e.g., fitness sharing and non-dominated sorting) because lexicase \textit{is} elitist for individual traits appearing early in the `shuffled' order of training cases used for selecting a parent.

To our knowledge, this is the first assessment of down-sampled lexicase on the exploitation rate diagnostic.
Our results show that random down-sampling (with no estimation) can substantially improve lexicase selection's ability to exploit smooth gradients. 
We attribute this improvement to down-sampled lexicase's use of fewer training cases (traits) when selecting a parent, which concentrates selection on candidate solutions that are elite on those cases. 
In search spaces where substantial exploration is required, selecting a large number of different parents can be beneficial.
However, the exploitation-rate diagnostic requires no off-gradient exploration. 
Concentrating selection to fewer individuals results in more focused mutations, increasing the probability of any single parent producing an offspring with beneficial mutations. 
Indeed, we found that standard lexicase selected, on average, close to 50\% of the population as parents (approximately 216,~\citep{supp_gh_repo}). 
Down-sampled lexicase (with no estimation) selected many fewer parents on average: approximately 120 and 28 for 1\% and 10\% subsampling levels, respectively.
All treatments with phylogeny-informed fitness estimation selected  approximately the same number of distinct parents as standard lexicase selection, providing some explanation for their under-performance on this diagnostic. 
That is, the diversity maintenance recovered by phylogeny-informed subsampling impeded exploitation rate. 

Preliminary evidence also suggests that lexicase selection is not robust to misleading performance information \citep{dolsonApplyingEcologicalPrinciples2018}.
As such, we also expect that inaccurate phylogeny-based fitness estimates could be contributing to slower exploitation rates. 

\subsubsection{Phylogeny-informed subsampling improves tournament selection's gradient exploitation rate}

\begin{figure}[h]
\begin{center}
\includegraphics[width=\linewidth]{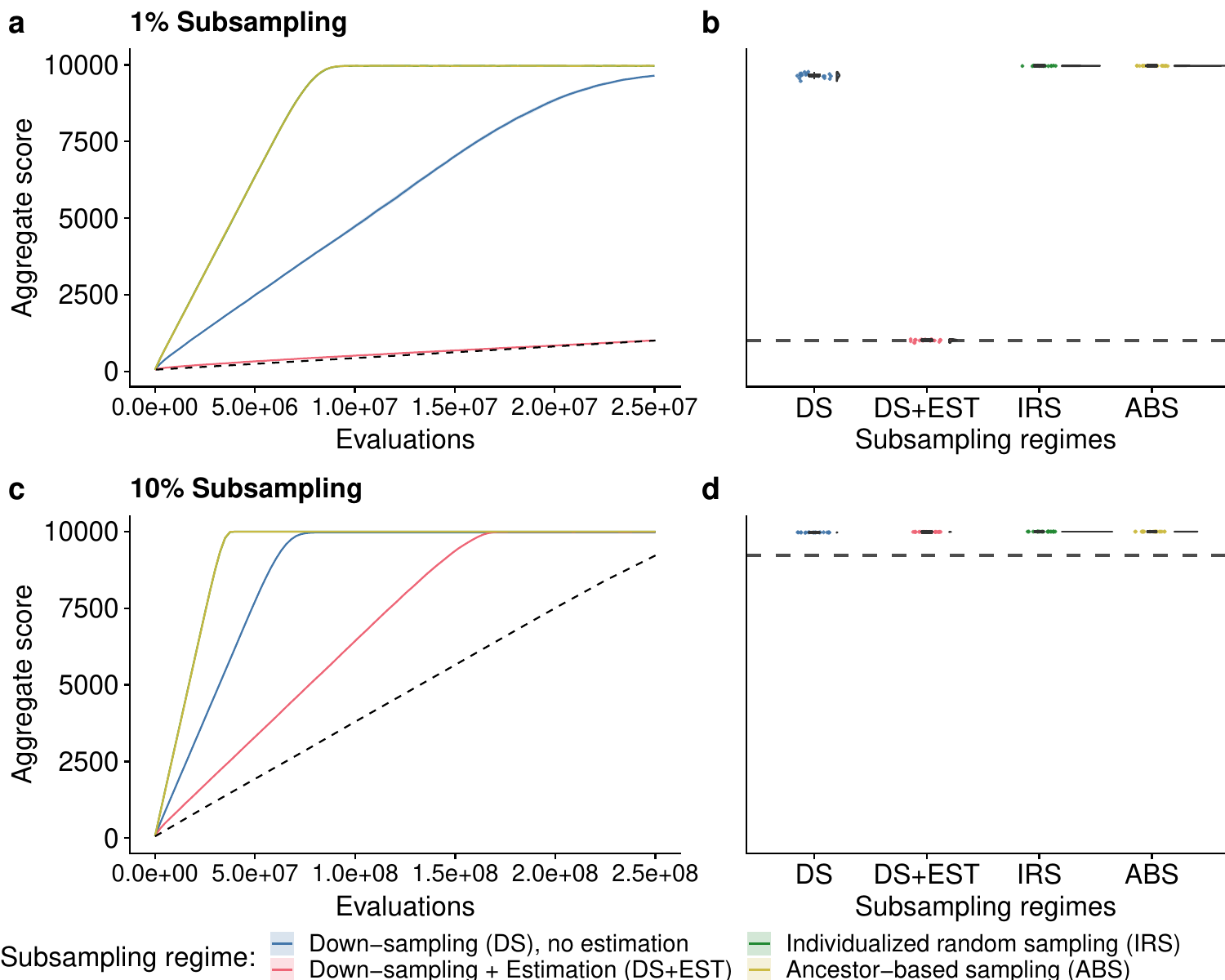}
\caption{\small{
\textbf{Aggregate trait score on the exploitation-rate diagnostic for subsampling regimes with tournament selection.}
Panels (a) and (c) show mean aggregate trait score over time for 1\% and 10\% subsampling levels, respectively.
Shading around the mean indicates a 95\% confidence interval. 
Panels (b) and (d) show best aggregate trait score after 50,000 generations of evolution for 1\% and 10\% subsampling levels, respectively.  
Dotted black lines in each plot indicate the median aggregate trait score achieved by standard tournament selection (no subsampling) after an equivalent number of trait evaluations.
Kruskal-Wallis tests for both subsampling levels were statistically significant ($p < 0.001$). 
}}
\label{fig:exploit-tourn}
\end{center}
\end{figure}

Previous work has shown that tournament selection with a sufficiently large tournament size is one of the best selection methods when pure exploitation is most important~\citep{hernandez_suite_2022}.
We wanted to assess the effect of phylogeny-informed subsampling in the context of a more exploitative selection scheme than lexicase, so we repeated our exploitation rate experiment using tournament selection (with a tournament size of 8).
Figure~\ref{fig:exploit-tourn} shows aggregate scores over time and after 50,000 generations of evolution under both 1\% and 10\% subsampling rates.

In contrast to our lexicase selection experiments, phylogeny-informed subsampling methods (IRS and ABS) outperformed all other subsampling regimes when combined with tournament selection (corrected Wilcoxon rank-sum tests, $p < 0.001$). 
At 1\% subsampling, all subsampling regimes except DS+EST achieved near-optimal scores for this diagnostic (Fig.~\ref{fig:exploit-tourn}b).
The DS+EST regime performed approximately the same as tournament selection without subsampling (for an equivalent number of evaluations). 
At 10\% subsampling, all subsampling regimes achieved near-optimal scores that were above the median score achieved in our standard tournament selection reference (Fig.~\ref{fig:exploit-tourn}d).
However, at both subsampling levels, Figures~\ref{fig:exploit-tourn}a and~\ref{fig:exploit-tourn}b reveal that IRS and ABS optimized populations at much faster rates than all other subsampling regimes.

Taken together with our lexicase selection results, our data show that choice of selection method determines which subsampling regimes perform best. 
For both selection methods, DS+EST performed worst. 
DS and DS+EST regimes evaluate all individuals in the population on the same training cases each generation.
When evaluating a population, IRS and ABS construct samples per-individual, likely resulting in the use of many more distinct training cases in a single generation than DS or DS+EST, which construct a single sample to use for the entire population.
As such, we hypothesize that fitness estimations may be less useful (or less accurate) under DS+EST versus the IRS or ABS subsampling regimes.

\subsection{Phylogeny-informed subsampling improves diversity maintenance and search space exploration}

\begin{figure}[h]
\begin{center}
\includegraphics[width=\linewidth]{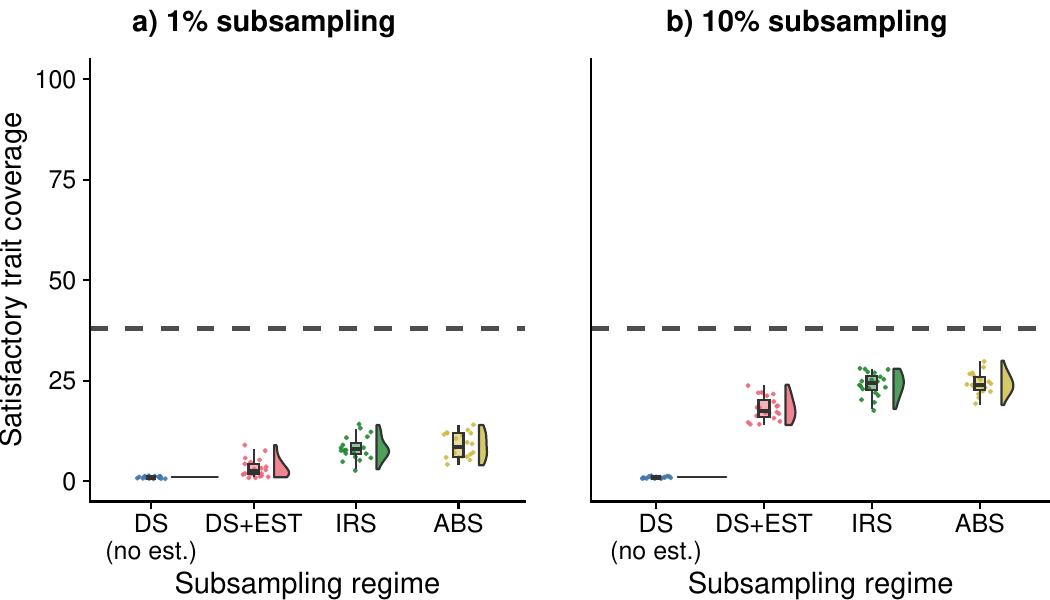}
\caption{\small{
\textbf{Satisfactory trait coverage on the contradictory objectives diagnostic.} 
Dotted black lines in both plots indicate the median aggregate trait score achieved by standard lexicase selection (no subsampling) after an equivalent number of trait evaluations.
Kruskal-Wallis tests for both subsampling levels were statistically significant ($p < 0.001$). 
}}
\label{fig:con-obj-lex}
\end{center}
\end{figure}

The contradictory objectives and multi-path exploration diagnostics both measure a selection method's ability to maintain diversity in a population. 
The contradictory objectives diagnostic isolates a selection scheme's ability to maintain many global optima in a population.
Consistent success on the multi-path exploration diagnostic requires that a selection scheme be capable of exploring many fitness gradients in the search space.

On the contradictory objectives diagnostic (Figure~\ref{fig:con-obj-lex}), no subsampling regime (at any subsampling level) matched or exceeded the trait coverage maintained by our standard lexicase reference.
Random down-sampling with no fitness estimation achieved the worst satisfactory trait coverage (corrected Wilcoxon rank-sum test, $p < 0.001$), maintaining just one optimal trait in all replicates for both 1\% and 10\% subsampling levels. 
Phylogeny-informed subsampling regimes, IRS and ABS, achieved the greatest satisfactory trait coverage at all subsampling levels relative to other subsampling regimes ($p < 0.001$). 
DS+EST averaged between 3 and 4 satisfactory traits at 1\% subsampling and between 18 and 19 at 10\% subsampling. 
Both IRS and ABS averaged between 8 and 9 and between 24 and 25 satisfactory traits at 1\% and 10\% subsampling, respectively.   

\begin{figure}[h]
\begin{center}
\includegraphics[width=\linewidth]{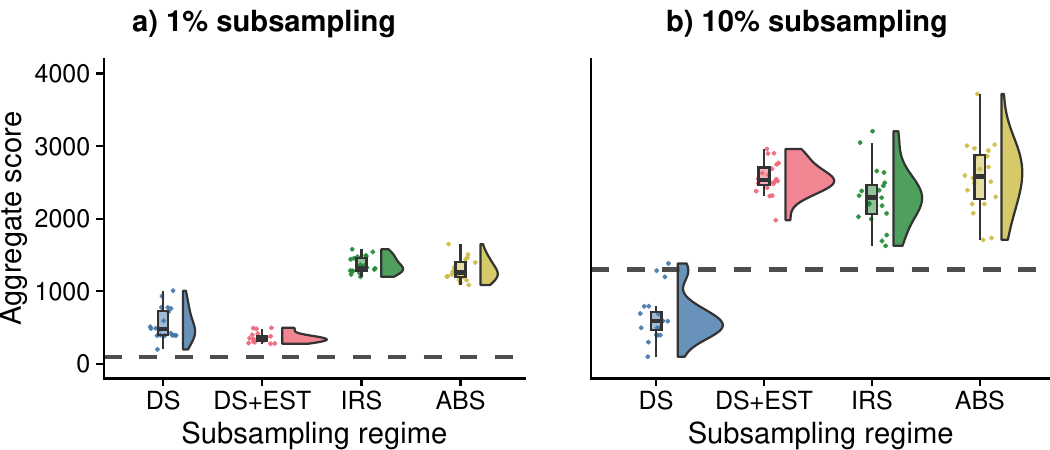}
\caption{\small{
\textbf{Aggregate trait score on the multi-path exploration diagnostic.} 
Dotted black lines in both plots indicate the median aggregate trait score achieved by standard lexicase selection (no subsampling) after an equivalent number of trait evaluations.
Kruskal-Wallis tests for both subsampling levels were statistically significant ($p < 0.001$). 
}}
\label{fig:explore-lex}
\end{center}
\end{figure}

On the multi-path exploration diagnostic at 1\% subsampling (Figure~\ref{fig:explore-lex}a), phylogeny-informed subsampling regimes (IRS and ABS) achieved the greatest aggregate trait scores (corrected Wilcoxon rank-sum test, $p < 0.001$), and we were unable to detect a significant difference between the IRS and ABS regimes.
The DS+EST regime achieved the worst aggregate score at 1\% subsampling ($p < 0.003$). 
All 1\% subsampling regimes achieved greater scores than standard lexicase run for an equivalent number of \textit{evaluations}. 

At 10\% subsampling (Figure~\ref{fig:explore-lex}b), random down-sampling without estimation (DS) achieved the worst performance of all subsampling regimes on the multi-path exploration diagnostic ($p < 0.001$). 
The DS+EST, IRS, ABS regimes each scored higher than our standard lexicase reference at an equivalent number of trait evaluations. 
We were unable to detect a significant difference among the DS+EST, IRS, and ABS regimes at 10\% subsampling.  

Our data from the contradictory objectives diagnostic are consistent with findings in a previous study, showing that phylogeny-based fitness estimation can improve diversity maintenance over random down-sampling with no estimation~\citep{lalejini_phylo_est_2023}. 
Indeed, naive random down-sampling can prevent lexicase selection from maintaining diverse populations~\citep{hernandez_exploration_2022,geiger_down-sampled_2023}. 
No subsampling regime fully recovered lexicase's capacity for diversity maintenance when using all training cases (traits) for parent selection.
However, our phylogeny-informed subsampling regimes did improve contradictory objective maintenance over down-sampling with fitness estimation.

Our data from the exploration diagnostic are consistent with previous work showing that phylogeny-based fitness estimation can improve diagnostic performance at sufficiently high subsampling levels but decrease performance at 1\% subsampling~\citep{lalejini_phylo_est_2023}. 
Data from the exploitation rate diagnostic offer an explanation for these results. 
At 1\% subsampling levels, DS+EST lexicase regimes fail to rapidly exploit fitness gradients (Figure~\ref{fig:exploit-lex}a). 
In order to explore many pathways in the multi-path exploration diagnostic, the population must be able to efficiently exploit each pathways fitness gradient.
Overall, phylogeny-informed subsampling regimes more consistently improved search space exploration as compared to using random down-sampling (with or without estimation). 

\subsection{Phylogeny-informed subsampling enables extreme subsampling on test-based problems}

\begin{table}[ht]
    \caption{\small{
    Problem-solving success on 10 program synthesis problems at 1\% and 10\% subsampling levels.
    The greatest success rate for each problem at each subsampling level is bolded. 
    Success rates that were significantly greater than the success rates of \textit{all other} regimes at that subsampling level are underlined (Holm-Bonferroni-corrected pairwise Fisher's exact tests, $p < 0.05$). 
    }}
    \centering
    \label{tab:psynth-success}
    \includegraphics[width=\linewidth]{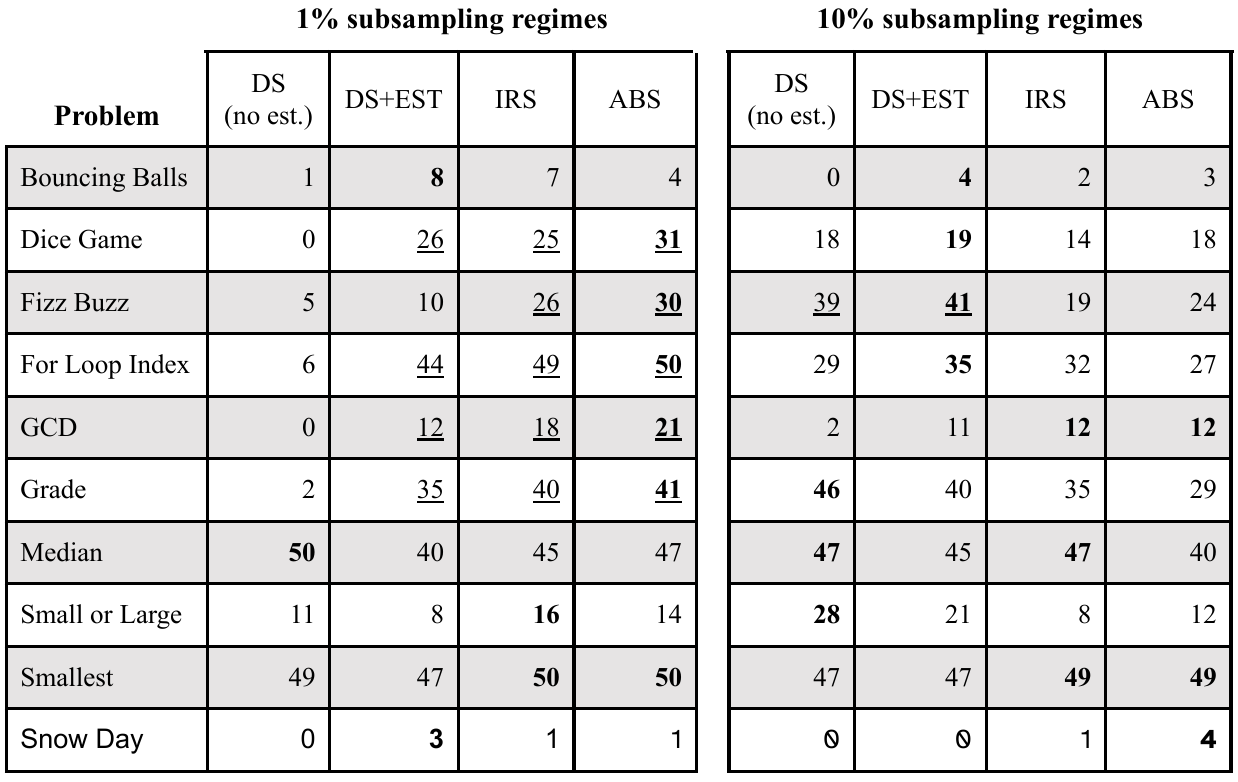}
\end{table}

Table~\ref{tab:psynth-success} shows problem-solving success on ten program synthesis problems from the general program synthesis benchmark suites~\citep{helmuth_general_2015,helmuth_psb2_2021}. 
We consider a run to be successful if it produces a program capable of passing all test cases in the testing set. 
For each problem, we compared the problem-solving success of each subsampling regime within a particular subsampling level.
We do not include a standard lexicase reference, but several studies have down that down-sampled lexicase typically outperforms standard lexicase for these problems~\citep{helmuth_benchmarking_2020,helmuthProblemSolvingBenefitsDownSampled2022}.

At 1\% subsampling, phylogeny-informed subsampling regimes (IRS or ABS) achieved the greatest success rates on 7 of 10 problems, 5 of which were significantly better than down-sampling without estimation (corrected pairwise Fisher's exact tests, $p < 0.05$).
We did not detect a significant difference between IRS and ABS on any problems. 
IRS and ABS significantly outperformed DS+EST on one problem (Fizz Buzz), and we were unable to detect differences between IRS, ABS, and DS+EST for all other problems at 1\% subsampling.  
Down-sampling with no estimation failed to produce solutions on three problems (Dice Game, GCD, and Snow Day) at 1\% subsampling. 
In all three of these cases, all three other subsampling regimes produced at least one solution.

At 10\% subsampling, phylogeny-informed subsampling regimes (IRS or ABS) produced the greatest success rates on 4 of 10 problems; none of these instances were significantly better than the success rates of all other regimes (DS and DS+EST).
DS produced the greatest success rates on 3 of 10 problems, and DS+EST produced the greatest success rates on 4 of 10 problems. 
On the Fizz Buzz problem, DS and DS+EST success rates were significantly better than both IRS and ABS ($p < 0.02$). 
Only DS and DS+EST failed to produce solutions on any problems; IRS and ABS runs produced solutions in at least one replicate for all problems. 

No single subsampling regime produced the greatest success rates across all problems and subsampling levels, which is consistent with other studies comparing selection methods on large problem sets~\citep{helmuth_benchmarking_2020,orzechowski_where_2018,jiang_evolutionary_2017}.
The particulars of the search space and genetic representation interact with the chosen evaluation and selection procedures when steering the population through a search space. 
At 1\% subsampling, however, our data suggest that regimes with some form of phylogeny-based fitness estimation performed best, and phylogeny-informed subsampling regimes are generally an improvement over down-sampling with estimation (Figure~\ref{tab:psynth-success}).

Successes at 1\% subsampling levels are impressive considering that each program in the population is evaluated on a single training case per generation.
At 1\% subsampling, random down-sampling with no estimation is reduced to elitist selection on a random training case.
This removes lexicase's ability to maintain multiple specialists on different subsets of the training cases~\citep{fergusonCharacterizingEffectsRandom2020}, which is known to be an important characteristic for lexicase's success~\citep{helmuth_lexicase_2019}.
Indeed, previous studies have shown that using too few training cases with standard or down-sampled lexicase reduces problem-solving success~\citep{schweim_effects_2022,fergusonCharacterizingEffectsRandom2020}.
We attribute much of the success of the DS+EST, IRS, and ABS regimes to their ability to use the full training set for selection, estimating a program's performance on any unevaluated training cases using that program's ancestry.
As such, these regimes can still maintain specialists by selecting a more diverse set of parents, and indeed, we found that DS+EST, IRS, and ABS regimes selected a greater diversity of parents than the DS regime~\citep{supp_gh_repo}. 
In supplemental analyses, we also found that phylogeny-informed subsampling methods generally had greater estimation accuracy than when estimating under random down-sampling, and estimation accuracy varied by problem.
For example, estimation accuracy was lowest for the Bouncing Balls and GCD problems, suggesting that mutations are potentially more likely to have phenotypic effects (making offspring more likely to be different from their parent) on these problems than on other problems.
Our results also suggest that the search spaces instantiated by the Median and Smallest problems are  trivial to navigate with simple hill climbing, and indeed, estimation accuracy was generally highest for these two problems. %

Our results at 10\% subsampling were far more mixed than at 1\% subsampling. 
The best choice of subsampling regime depended on the particular problem. 
At 10\% subsampling, each program is evaluated against 10 training cases per generation.
Training sets in program synthesis typically highly redundant; that is, there are many training cases that assess the same mode of behavior.
If sampling just one training case (1\% subsampling), it is impossible to sample a representative set of training cases for most problems (e.g., those that require programs to produce qualitatively different outputs for different inputs). 
However, with 10 training cases (10\% subsampling), it is much more likely to capture a representative sample of training cases without omitting cases that assess a particular mode of behavior~\citep{spectorAssessmentProblemModality2012}. 

While some problems clearly benefited from the use of phylogenetic analysis to aid evaluation and selection at 10\% subsampling (e.g., Bouncing Balls, Snow Day, and GCD), most other problems did not.
This suggests that the forms of phylogeny-based estimation and phylogeny-informed subsampling are most appropriate for enabling \textit{extreme} subsampling, which is often necessary for domains where evaluation is particularly expensive (e.g., any that require expensive simulations to evaluate candidate solutions~\citep{moore_limits_2019}).

\section{Conclusion}

In this work, we demonstrated two simple phylogeny-informed subsampling methods: individualized random subsampling (IRS) and ancestor-based subsampling (ABS). 
Relative to random down-sampling, we found evidence that phylogeny-based subsampling methods can reduce lexicase selection's ability to exploit a smooth fitness gradient (Fig.~\ref{fig:exploit-lex}), but improve tournament selection's ability to exploit gradients (Fig.~\ref{fig:exploit-tourn}).
We also found that phylogeny-informed subsampling methods can improve diversity maintenance and overall search space exploration (Fig.~\ref{fig:con-obj-lex} and~\ref{fig:explore-lex}).
Across ten program synthesis benchmark problems, we found that subsampling methods that take advantage of phylogenetic analysis allow for problem-solving success at extreme subsampling levels where random down-sampling often fails (Fig.~\ref{tab:psynth-success}).
However, at more moderate subsampling levels, phylogeny-informed subsampling generally performed no better than random subsampling on program synthesis problems.

Future studies are necessary to further characterize these failure cases for phylogeny-informed subsampling, which will inform further refinements to these techniques. 
For example, inaccurate fitness estimates likely reduce problem-solving success. 
One potential source of inaccuracies are scenarios where offspring have large mutations. 
Phylogenies could be additionally annotated with mutation information, and we could take mutational differences into account when estimating fitnesses and constructing subsamples (e.g., as in~\cite{pilato_fitness_2007}).  

Overall, phylogeny-informed methods show promise for helping to scale up evolutionary computing systems to tackle problems with many fitness criteria that are each expensive to evaluate.
In particular, we recommend phylogeny-informed methods on problems where evaluating a large population on many fitness criteria is not computationally feasible, necessitating extreme subsampling.

\begin{acks}

We thank our respective universities for supporting this work.
This research was supported by Grand Valley State University through a Kindschi Fellowship to MS.
Additionally, this work was supported in part through computational resources and services provided by the Institute for Cyber-Enabled Research at Michigan State University.

\end{acks}

\bibliographystyle{ACM-Reference-Format}
\bibliography{references,supplemental}

\end{document}